\documentclass{article}
\usepackage{style_file}

\usepackage{times}
\usepackage{soul}
\usepackage{url}
\usepackage[hidelinks]{hyperref}
\usepackage[utf8]{inputenc}
\usepackage[small]{caption}
\usepackage{graphicx}
\usepackage{amsmath}
\usepackage{amsthm}
\usepackage{booktabs}
\usepackage{algorithm}
\usepackage{algorithmic}
\usepackage[switch]{lineno}

\usepackage{amsfonts}
\usepackage{amssymb}
\usepackage{latexsym}
\usepackage{caption}
\usepackage{tabto}
\usepackage{upgreek}
\usepackage{lipsum}
\DeclareSymbolFont{cmbrightop}{OT1}{cmbr}{m}{n}
\DeclareMathSymbol{\sfPsi}{\mathalpha}{cmbrightop}{9}
\newcommand{\mymath}[1]{\begin{math}#1\end{math}}
\newcommand\blfootnote[1]{%
  \begingroup
  \renewcommand\thefootnote{}\footnote{#1}%
  \addtocounter{footnote}{-1}%
  \endgroup
}

\title{Img2Tab: Automatic Class Relevant Concept Discovery from StyleGAN Features for Explainable Image Classification}

    \author{
    Youngjae Song$^1$\and
    Sung Kuk Shyn$^2$\and
    Kwang-su Kim$^{3}$\footnotemark
    \affiliations
    $^1$Department of Computer Science and Engineering, Sungkyunkwan University, South Korea\\
    $^2$Department of Artificial Intelligence, Sungkyunkwan University, South Korea\\
    $^3$Department of Computing and Informatics, Sungkyunkwan University, South Korea\\
    \emails
    \{yuong13, davidshyn, kim.kwangsu\}@skku.edu\\
    }

\begin{document}
\twocolumn[{
\renewcommand\twocolumn[1][]{#1} 
 \maketitle
\begin{center}
    \centering
    \captionsetup{type=figure}
    \includegraphics[width=0.9\textwidth]{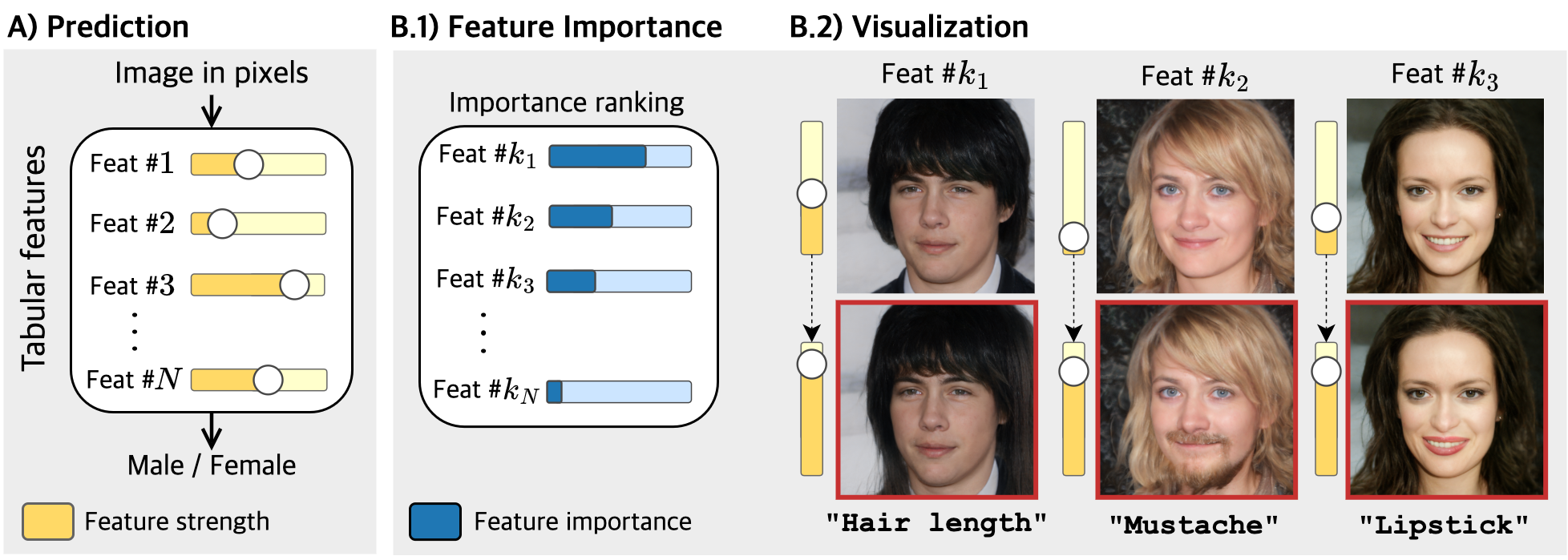}
    \caption{\textbf{An outline of an Img2Tab example in a gender classification.} A) Img2Tab predicts its label of the image by tabular features encoded from the image, not by pixels. B) Img2Tab explains its prediction by feature importance and visualization at the concept level. We visualize concepts in the red boxes by latent modifications. Important features (\#\mymath{k_1}, \#\mymath{k_2}, and \#\mymath{k_3}) extracted by the tabular classifier represent class-relevant concepts (``Hair length," ``Mustache," and ``Lipstick," respectively).}
\end{center}
}]

\blfootnote{*Corresponding author}

\begin{abstract}
\quad Traditional tabular classifiers provide explainable decision-making with interpretable features(\textit{concepts}). However, using their explainability in vision tasks has been limited due to the pixel representation of images. In this paper, we design Img2Tabs that classify images by concepts to harness the explainability of tabular classifiers. Img2Tabs encode image pixels into tabular features by StyleGAN inversion. Since not all of the resulting features are class-relevant or interpretable due to their generative nature, we expect Img2Tab classifiers to discover class-relevant concepts automatically from the StyleGAN features. Thus, we propose a novel method using the Wasserstein-1 metric to quantify class-relevancy and interpretability simultaneously. Using this method, we investigate whether important features extracted by tabular classifiers are class-relevant concepts. Consequently, we determine the most effective classifier for Img2Tabs in terms of discovering class-relevant concepts automatically from StyleGAN features. In evaluations, we demonstrate concept-based explanations through importance and visualization. Img2Tab achieves top-1 accuracy that is on par with CNN classifiers and deep feature learning baselines. Additionally, we show that users can easily debug Img2Tab classifiers at the concept level to ensure unbiased and fair decision-making without sacrificing accuracy.
\end{abstract}

\section{Introduction}
\quad The performance of image classifications has soared markedly, but high performance is still insufficient to deploy AI products in computer vision domains. Many researchers deem XAI(eXplainable AI) crucial to design trustworthy, unbiased, and fair AI models. Despite recent efforts, human understanding of the intrinsic decision-making process in deep net classifiers proves elusive. Alternatively, traditional tabular classifiers provide inherently explainable decision-making of high-level features. Nevertheless, pixel representation of images hinders their explainability from being applied to vision tasks. Thus, we explore an alternative framework rather than solely relying on deep net classifiers at the pixel level. 

Generally, tabular features are assigned in quantitative semantics as discrete or continuous scalars. Also, the relationship between the feature scalar and semantics strength is monotone; a total order of feature scalar corresponds to a total order of semantics strength. These tabular properties allow us to understand each feature value with only one dimension. Meanwhile, tabular models such as Logistic Regression \cite{berkson1944application}, SVM with linear kernels \cite{hearst1998support}, XGBoost \cite{chen2016xgboost}, and TabNet \cite{arik2021tabnet} are designed to make explainable decisions. According to \cite{borisov2021deep}, explainability is becoming essential for tabular models as a useful tool for debugging and auditing models' decision-making. This paper argues that StyleGAN features can be regarded as tabular data; we investigate which tabular models are suitable for using such features as input data to accomplish explainable, concept-based image classification.

In Section 3, we analyze StyleGAN \cite{karras2020analyzing} feature properties to implement them as tabular features. As discussed in \cite{nitzan2022large}, distances along the StyleGAN latent are globally consistent: a scalar distance along the latent element represents its semantic strength. As such distance is monotonic, the StyleGAN feature values apply to tabular feature values. Given the advancement of GANs, we exploit StyleGAN feature space to encode image pixels into tabular features.

The StyleGAN feature space represents \textit{generative} features indicating that not all features are class-relevant. \cite{wu2021stylespace} demonstrates that not all StyleGAN features are concepts. For the sake of explainable image classifications, we expect our classifier to discover class-relevant concepts automatically from such a feature space. To achieve this goal, we propose a novel method using the Wasserstein-1 metric \cite{kantorovich1960mathematical} to measure both class-relevancy and interpretability of given StyleGAN features. The metric indicates that important features extracted by XGBoost \cite{chen2016xgboost} are the most class-relevant and interpretable compared to other tabular classifiers.

Based on the previous findings, we propose \textbf{Img2Tab}, a framework that classifies images and explains its predictions based on automatically discovered class-relevant concepts. By implementing GAN inversion and tabular classifier, Img2Tabs consist of three modules sharing the StyleGAN feature space: an encoder, a generator, and a classifier. They cooperate to predict, explain, and debug at the concept level. Img2Tabs provide explanations by importance and visualization at the concept level. By harnessing the explainability of tabular models, Img2Tabs identify the importance of each tabular feature. The StyleGAN generator visualizes concept semantics by \textit{latent modification}. By altering the concepts' value of interest, their semantics resolve into generated images as visual displays for users.

It is to be emphasized that Img2Tabs serve not to explain the pre-trained classifiers but rather to design an inherently explainable framework. Unfortunately, inherently explainable models usually struggle with relatively lower performance\cite{molnar2018guide} than black-box classifiers. The key to designing practical and interpretable models involves reaching comparable performance as black-box models. Section 5 exhibits that Img2Tab achieves top-1 accuracy, on par with black-box baselines.

\textit{Fairness} has evolved as a core part of designing AI models, which should rely on something other than biased or unfair concepts for their decision-making. However, it is challenging to filter specific semantics out of the image. Masking a particular pixel region in the image may suffice, but this unnecessarily removes all information in the region. Because of the concept-level decision-making of Img2Tabs, users can easily debug the Img2Tabs classifier by masking specific unwanted concepts before training. In Section 5, we evaluate how debugging affects the top-1 accuracy and concept importance.

To our best knowledge, Img2Tabs become the first inherently explainable framework that classifies high-resolution images based on human-understandable concepts in an unsupervised manner, while providing explanations through importance and with clear visualization at the concept level. Most importantly, it encounters no reduction in accuracy. The key contributions of this paper are summarized as follows:
\begin{itemize}
    \item We propose a novel method using the Wasserstein-1 metric to quantify the class-relevancy and interpretability of given StyleGAN features simultaneously.
    \item We design an Img2Tab framework that classifies images and explains prediction based on automatically discovered, class-relevant concepts. 
    \item Img2Tab achieves top-1 accuracy on par with CNN and deep feature learning baselines.
    \item Users can easily debug Img2Tab classifier at the concept level to ensure unbiased and fair decision-making. 
\end{itemize}

\section{Related Work}
\quad \textbf{Latent Space of GANs} Many works leverage the advancement of GANs for feature learning. For example, StyleGAN has attracted considerable attention due to its well-disentangled, smooth, and semantically rich feature space. Recent works such as Styleflow \cite{abdal2021styleflow}, GANspace \cite{harkonen2020ganspace}, InterfaceGAN \cite{shen2020interfacegan}, and StyleSpace \cite{wu2021stylespace} presented techniques to discover the concepts encoded in the GAN latent. Such concepts appear semantically linear on the generated images when interpolated along a certain latent direction. To edit real images along the discovered concepts, users should reconstruct the images via GAN generators. This idea has accelerated the advent of GAN inversion methods for reconstructing images; such methods include ALI \cite{dumoulin2016adversarially}, e4e \cite{tov2021designing}, pSp \cite{richardson2021encoding}, HyperStyle \cite{alaluf2022hyperstyle}, ReStyle \cite{alaluf2021restyle}, and PTI \cite{roich2022pivotal}). GAN inversion has motivated another possibility as a feature learning method for downstream tasks. Several works(LARGE \cite{nitzan2022large}, GHFeat \cite{xu2021generative}, BiGAN \cite{donahue2016adversarial}, BigBiGAN \cite{donahue2019large}, and ALI \cite{dumoulin2016adversarially}) exploit the features projected from real images for regression or classification tasks. On account of the interpretability of StyleGAN features, we extensively leverage them in the view of XAI. Section 5 compares Img2Tabs performance with deep feature learning methods on classification accuracy.

\textbf{Tabular classifiers} Tabular classifiers such as 
 Linear Regression \cite{berkson1944application}, SVM \cite{hearst1998support}, XGBoost \cite{chen2016xgboost}, CatBoost \cite{dorogush2018catboost}, and TabNet \cite{arik2021tabnet} are widely used in tabular classification tasks due to their inherent interpretability. They provide an explanation of feature importance, one of the more intuitive methods. The explanation enables users to understand its inference and to debug its decision-making at the human-labeled feature level.

\textbf{Concept-based Explanations} A main challenge for concept-based explanations is to extract high-level concepts and visualize them. CBM \cite{koh2020concept}, TCAV \cite{kim2018interpretability}, and CoCoX \cite{akula2020cocox} require concept annotation or additional information about concepts. SENN \cite{Alvarez2018towards}, ConceptSHAP \cite{yeh2020completeness}, and \cite{sarkar2022framework} visualize concepts by showing samples with high concept activation. ACE \cite{ghorbani2019towards} and CoCoX \cite{akula2020cocox} visualize concepts in super-pixels, which are unclear. In the case of DISSECT \cite{ghandeharioun2021dissect}, EPE \cite{singla2019explanation}, SENN \cite{Alvarez2018towards}, and ConceptSHAP \cite{yeh2020completeness}, they fail to provide explanations in high-resolution images. Lastly, those who learn concepts by jointly training GAN(DISSECT \cite{ghandeharioun2021dissect}, EPE \cite{singla2019explanation}, and StylEx \cite{lang2021explaining}) ought to train the whole framework(including the generator) as they cannot use pre-trained generators.

\section{Setup for Designing Img2Tabs}
\quad In this section, we argue why StyleGAN features are effectively tabular features. By studying the feature properties from the XAI perspectives, we present a novel method using the Wasserstein-1 metric to simultaneously measure the class-relevancy and interpretability of given StyleGAN features. Then, we exploit the method to determine which tabular model proves the most effective at automatically discovering class-relevant concepts without requiring any manual feature selection. 

\subsection{StyleGAN latent for tabular features}
\quad Let $\mathcal{X} \subset \mathbb{R}^{H \times W \times C}$ be the input data(image set), $\mathcal{Y} \subset \mathbb{R}$ be the output data(label set), where $x \in \mathcal{X}$ and $y \in \mathcal{Y}$. We define our dataset with $N$ samples as follows, $\mathcal{D} = \{(x_i,y_i)^N_{i=1} \}$. The StyleGAN generator $G:\sfPsi \rightarrow \mathcal{X}$ reconstructs the input image $x$ to $\hat{x}$, where $\sfPsi \subset \mathbb{R}^{d_\sfPsi}$ is a feature space containing concepts. We define a GAN encoder $E:\mathcal{X} \rightarrow \sfPsi$ transforming image $x$ into $\sfPsi$ space, $\psi = E(x)$, to encode image pixels into tabular features. Note that $\psi_{i,k}$ stands for the $i$-th sample's $k$-th feature value , which potentially could be a concept. In this exercise, we assume three conditions for a latent to be regarded as a tabular feature. First, feature values are assigned quantitatively in human-understandable semantics as discrete or continuous scalars. Second, each dimension in the latent represents a single semantic by itself. Third, the relationship between the feature scalar $\psi_{i,k}$ and semantic strength in the image $\hat{x_i}$ is monotonic within a certain range of values. This implies that directionally controlling the feature scalar $\psi_{i,k}$ reveals the corresponding semantic in the generated image $\hat{x}_{i}$ in the same direction.

We argue that a StyleGAN latent satisfies the three conditions for being tabular features. \mymath{\sfPsi} supports three latents \mymath{\mathcal{Z}}, \mymath{\mathcal{W}}, and \mymath{\mathcal{S}}. \mymath{\mathcal{Z}} denotes Gaussian prior, while we obtain the remaining latents by feeding them forward through a mapping network of StyleGAN (\mymath{\mathcal{W}}) and subsequent affine transformation layer (\mymath{\mathcal{S}}). Each latent spans in a continuous domain. In addition, as studied in previous research\cite{wu2021stylespace}, \mymath{\mathcal{S}} space exhibits improved disentanglement compared to other latent spaces \mymath{\mathcal{Z}} and \mymath{\mathcal{W}}. Improved disentanglement implies that each feature dimension represents a single, lone semantic. Moreover, \cite{wu2021stylespace} demonstrates that controlling a single element in $\mathcal{S}$ leads to a consistent and monotonic change in generated images \mymath{\hat{x_i}}. Thus, the properties of the space \mymath{\mathcal{S}} satisfy the three desirable conditions for being a tabular feature space. As a result, we leverage \mymath{\mathcal{S}} space as \mymath{\sfPsi} in our Img2Tab frameworks.

\subsection{Discovering class-relevant concepts within $\sfPsi$}
\quad An explainable classifier should make decisions by class-relevant concepts, which are interpretable and useful for classification. Since the GAN inversion networks \mymath{G(E(\cdot))} reconstruct a real image \mymath{x_i}, the latent space \mymath{\sfPsi} contains all image features, even those that are not useful for classification tasks. As researched by \cite{wu2021stylespace}, only a small portion of features in \mymath{\sfPsi} are interpretable. Thus, explainable classifiers must discover such features that are both class-relevant and interpretable from \mymath{\psi_i}. In this section, we present a novel method using the Wasserstein-1 metric in order to reveal quantitatively that an existing tabular classifier automatically discovers class-relevant concepts within the StyleGAN latents \mymath{\sfPsi}.

The encoder \mymath{E} encodes all samples $x_i$ from the train dataset \(\mathcal{D}_{train}\), and store each row vector $\psi_i \in \mathbb{R}^{d_{\sfPsi}}$ into a matrix $\Psi \in \mathbb{R}^{|\mathcal{D}_{train}|\times d_{\sfPsi}}$ as illustrated in Figure 2.

\begin{figure}[H]
  \centering
  \includegraphics[width=0.78\linewidth]{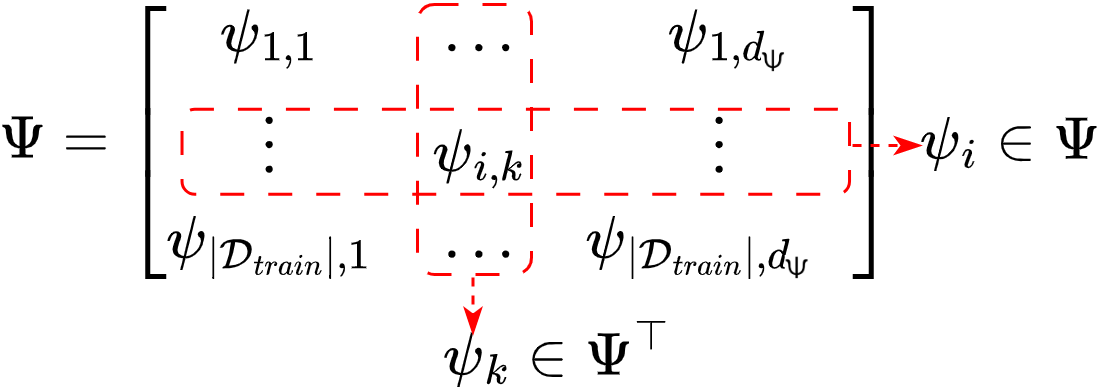}
  \caption{$\psi$ matrix illustration}
   \label{fig:onecol}
\end{figure}

\begin{figure*}[!t]
  \centering \includegraphics[width=1.0\linewidth]{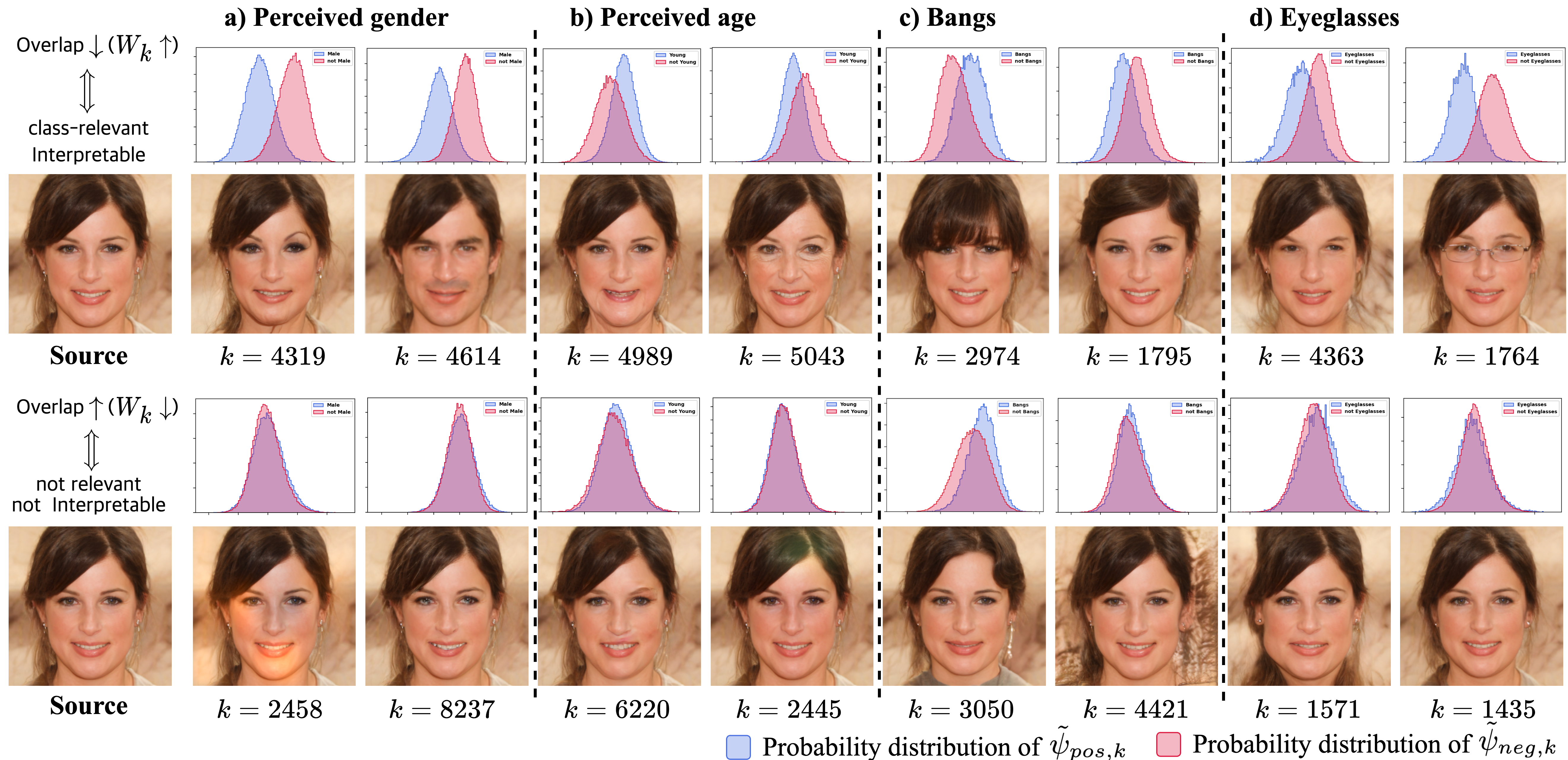}
  \caption{This figure depicts plots of probability distributions between $\Tilde{\psi}_{pos,k} \in \Tilde{\Psi}_{pos}^\top$(blue) and $\Tilde{\psi}_{neg,k} \in \Tilde{\Psi}_{neg}^\top$(red) with corresponding $\psi_{i,k}$ visualization.
  The upper half presents the results of top-2 $k$ with the largest $W_k$. Refer to Supplementary Figure 1 for more concepts. The lower half presents the results of randomly chosen $k$. We observe that semantics of $\psi_{k}$ appear strongly class-relevant and interpretable when the two probability distributions are less overlapped each other. On the other hand, randomly chosen features are neither interpretable nor relevant to the class. Note that $\psi_{i,3050}$ seems relevant to the class, a mere coincidence by random sampling. This figure highlights $W_k$ as an empirical metric, discovering class-relevant concepts.}
  \label{visina8}
\end{figure*}

\begin{algorithm}[t!]
    \caption{Measure \mymath{W_k} of given StyleGAN features}
    \label{alg:algorithm}
    \begin{algorithmic}[] 
        \STATE \textbf{Data}: Train dataset $\mathcal{D}_{train}$, Encoder \mymath{E}.
        \STATE \textbf{Result}: An array $W_k$ \\
        \STATE \textbf{Initialization}: $\Psi=$ 2-d matrix, $W=$ 1-d array \\
        \FOR{$(x_i, y_i)$ in $\mathcal{D}_{train}$}
            \STATE $\psi_i = E(x_i)$
            \STATE Add $\psi_i$ to $\Psi$ as a row vector
        \ENDFOR
        \FOR{$\psi_k$ in $\Psi^\top$}
            \STATE Standardize $\psi_k$
        \ENDFOR
        \STATE $\Tilde{\Psi}_{pos}=\{\Tilde{\psi}_i \in \Psi \ | \ y_i=1, \ (x_i,y_i) \in \mathcal{D}_{train} \}$
        \STATE $\Tilde{\Psi}_{neg}=\{\Tilde{\psi}_i \in \Psi \ | \ y_i=-1, \ (x_i,y_i) \in \mathcal{D}_{train} \}$
        \FOR{$(\Tilde{\psi}_{pos,k}, \Tilde{\psi}_{neg,k})$ in $(\Tilde{\Psi}_{pos}^\top, \Tilde{\Psi}_{neg}^\top)$}
            \STATE $W_k =$ Wasserstein-1($\Tilde{\psi}_{pos,k}$, $\Tilde{\psi}_{neg,k}$)
            \STATE Add $W_k$ to $W$
        \ENDFOR 
        \STATE \textbf{return} $W$
    \end{algorithmic}
\end{algorithm}

Then, we standardize each $k$-th column vector $\psi_k \in \Psi^\top$. Considering binary classification, we split the $\Tilde{\Psi}$ into positive samples $\Tilde{\Psi}_{pos}$ and negative samples $\Tilde{\Psi}_{neg}$ according to their labels $y_i$. Note that the tilde sign implies ``after standardization." We assume that the scalar $\Tilde{\psi}_{i,k}$ consistently conveys the strength of a certain semantic if $\psi_{i,k}$ represents an interpretable feature. The scalar $\Tilde{\psi}_{i,k}$ frequently fires distinguishably according to the label if the $\psi_{i,k}$ feature represents useful information for the classification. For example, if $\psi_{i,k}$ represents the concept of ``makeup," $\Tilde{\psi}_{i,k}$ scalars will tend to be frequently and consistently higher in female (as opposed to male) images. Specifically, probability distributions of each $k$-th feature across the data samples between $\Tilde{\psi}_k \in {\Tilde{\Psi}_{pos}^\top}$ and $\Tilde{\psi_k} \in \Tilde{\Psi}_{neg}^\top$ will be distinct, ergo their probability distribution plots do not overlap. The probability distributions will overlap if the feature $\psi_{i,k}$ does not satisfy either class-relevancy or interpretability. To quantify the degree of overlap, we compute the Wasserstein-1 distance $W_k$ between the two probability distributions. The detailed procedure is described in Algorithm 1.

\begin{figure}[!t]
  \centering
  \includegraphics[width=0.95\linewidth]{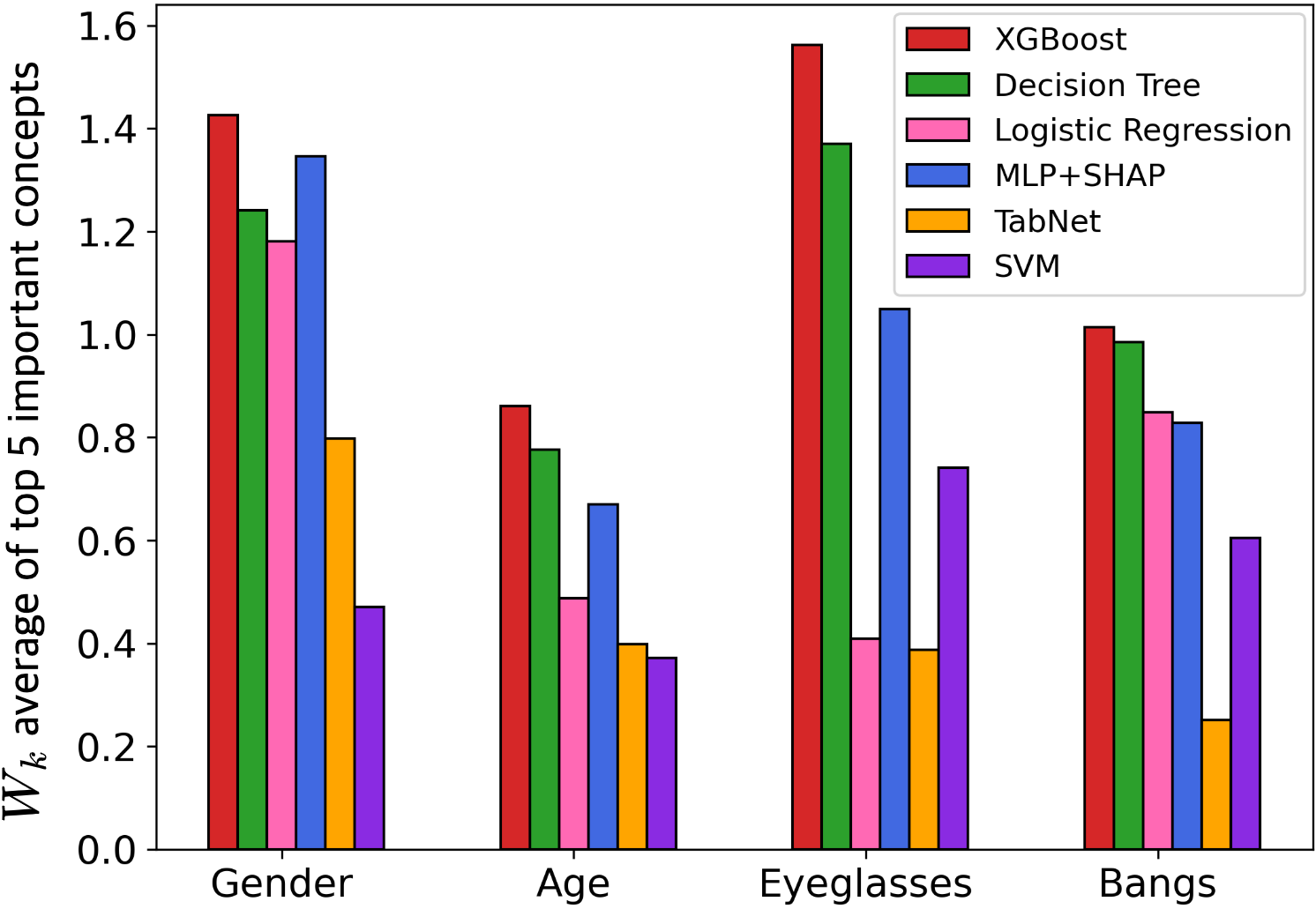}
  \caption{This figure compares the average $W_k$ of top-5 important features discovered by interpretable tabular classifiers for various classification tasks. In every task, XGBoost classifier outperforms all other models at \textit{automatically} discovering class-relevant concepts within $\sfPsi$, yielding the highest $W_k$. Note that we obtain globally important features of TabNet by counting the frequency of locally important features across all test samples.}
   \label{fig:onecol}
\end{figure}

\begin{figure*}[!t]
  \centering \includegraphics[width=1.0\linewidth]{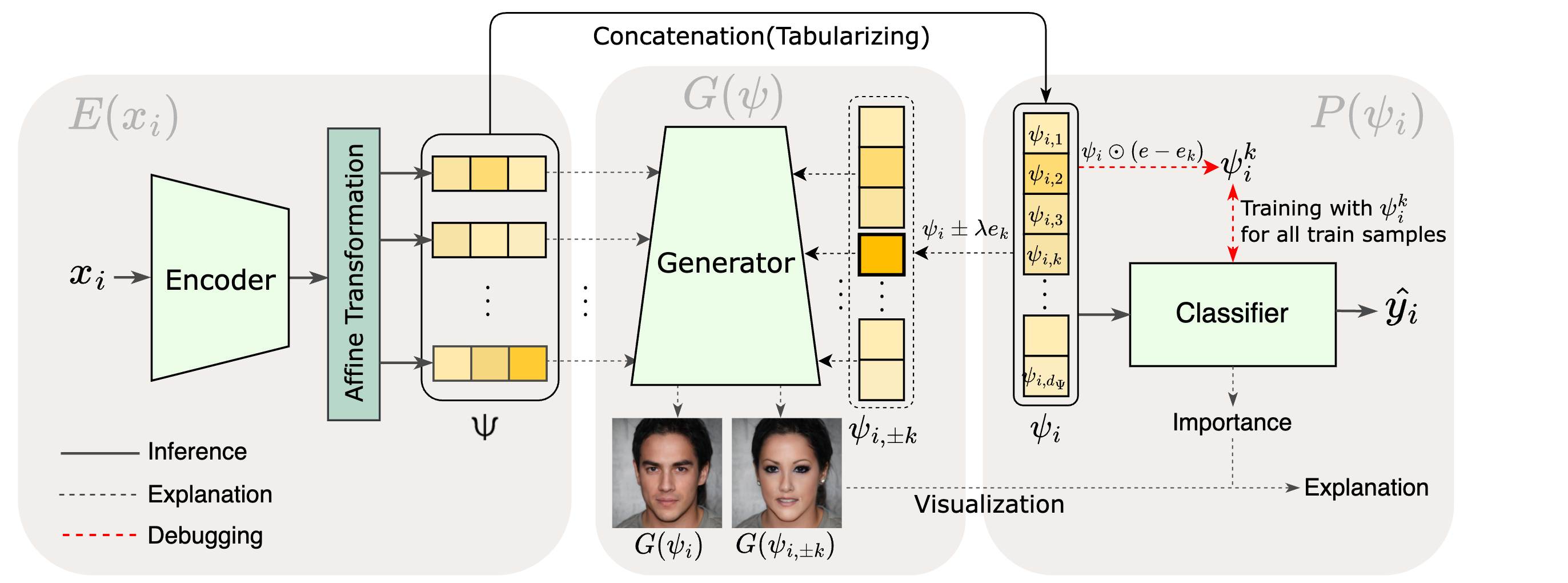}
  \caption{\textbf{Architecture of Img2Tab frameworks in details.} The image $x_i$ is transformed via $E$ into StyleGAN latent space $\sfPsi$. A user concatenates the extracted latent into the tabular form $\psi_i$. The classifier $P$ predicts the label $hat{y_i}$ from $\sfPsi$ at the concept level automatically discovering class-relevant concepts within $\psi_i$. In the explanation stage, the classifier provides us an explanation by importance and visualization $G(\psi_{i,\pm k})$. For debugging, users train the classifier $P$ with train samples of which unwanted concepts $\psi_{i,k}$ are masked into $\psi_i^{k}$. Note that we sequentially train each network in the order of $G$, $E$, and $P$, sharing the $\sfPsi$ space as a bottleneck.}
\end{figure*}

For empirical justifications, we search top-2 $\psi_{k}$ with the highest $W_k$. Figure 3 depicts distributions of top-2 $\psi_{k}$ for each class and concept visualization obtained by latent modifications. The less distributions overlap, the more semantics of the $\psi_{k}$ are both class-relevant and interpretable. Therefore, we use $W_k$ to measure whether or not the given StyleGAN features are class-relevant concepts.

We investigate whether important features extracted by existing tabular classifiers are class-relevant concepts. Figure 4 illustrates the $W_k$ average values for the top 5 important features discovered by each classifier. We train the classifiers with dataset of $\psi_{train}=\{E(x_i) \ | \ x_i \in \mathcal{D}_{train}\}$. Note that we employ SHAP\cite{lundberg2017unified} for MLP and a linear kernel for SVM to compute feature importance. The evaluation illustrates that important features extracted by the XGBoost classifier are the most class-relevant and interpretable. Therefore, we implement the XGBoost Img2Tab classifier as it automatically distinguishes class-relevant concepts by extracting important features.

\section{Img2Tab Frameworks}
\quad In this section, we design an Img2Tab framework based on the finding from Section 3.

\quad \textbf{Architecture and concept-based inference} As described in Figure 5, three main components incorporated in bottleneck architecture share the latent space $\sfPsi$: 1) a GAN encoder $E$ for the concept extractor, 2) a StyleGAN generator $G$ for the concept visualizer, and 3) a tabular classifier $P$ for classifying images based on concepts. The generator $G$ and the classifier $P$ cooperate in providing an explanation by importance and visualization at the concept level. We implement e4e \cite{tov2021designing} for the GAN inversion network $\hat{x_i}=G(E(x_i))$ as e4e is well-designed to manipulate and visualize semantics of specific features. To classify an image $x_i$ from extracted tabular feature vector $\psi_i$, we implement a tabular classifier $P:\sfPsi \rightarrow \mathcal{Y}$. As discussed in Section 3.2, we choose XGBoost for the classifier $P$. Img2Tab predicts the label of images in an end-to-end fashion: $\hat{y_i}=P(E(x_i))$.

\begin{figure*}[]
  \centering \includegraphics[width=1.0\linewidth]{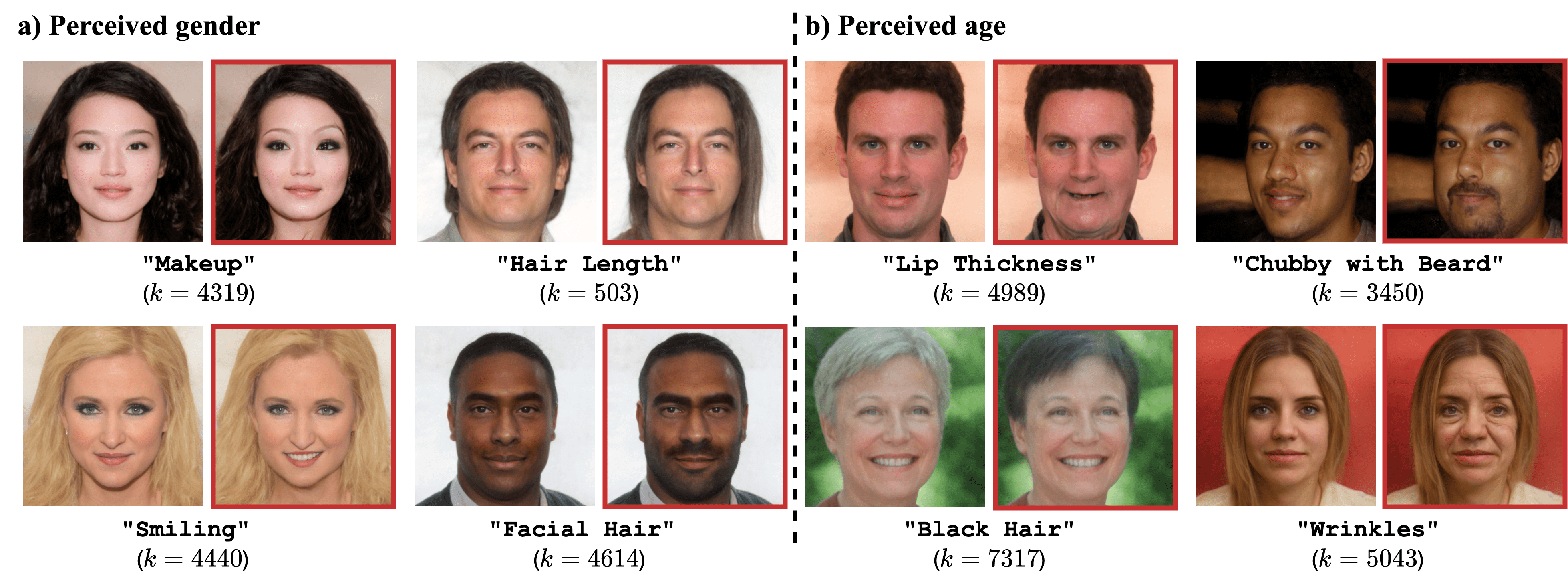}
  \caption{Top 4 important concepts automatically discovered by Img2Tab with the XGboost classifier. For each image pair, the left image signifies for the source image, and the one with the red box represents $G(x_{i,\pm k})$ of $\psi_{i,k}$ for the corresponding image. The important concepts represent diverse, well-disentangled, class-relevant, and human-understandable semantics.}
\end{figure*}
 
\textbf{Concept-based explanation} Img2Tab framework explains its prediction in two steps: it discovers important concepts and visualizes them. Then, generator $G$ visualizes the semantics of important concepts by latent modification. Many works of literature such as EPE \cite{singla2019explanation}, DISSECT \cite{ghandeharioun2021dissect}, and StylEx \cite{lang2021explaining}) exploit latent modification for visualizing concepts. In the same fashion, we redefine the latent modification used in our framework. A modified concept vector after modification in the latent level is defined as follows: $\psi_{i,\pm{k}} = \psi_i \pm \lambda e_k$, where $e_k \in  \mathbb{R}^{d_\psi}$ stands for a unit vector of which $k$-th element's value is 1. To be more precise, $\psi_{i,\pm{k}}$ depicts a concept vector of which $k$-th element is modified as much as $\lambda \in \mathbb{R}$ in either positive or negative direction. Then, we feed the modified concept vector into the generator $G$: $\hat{x}_{i,\pm k} = G(\psi_{i,\pm{k}})$. We expect $G(\psi_{i,\pm k})$ to manifest a change of the $k$-th concept's semantic. $G(\psi_i)$ and $G(\psi_{i,\pm k})$ together enable users to observe the semantics of the interesting concepts. After all, we identify important concepts in decision-making and which semantics it matches.

\quad \textbf{Concept-based debugging} Img2Tab enables users to identify class-relevant concepts either from the $W_k$ metric or classifier $P$. Once Users examine whether the class-relevant concept represents biased or unfair semantics, they may filter the semantics out of classifier $P$ decision-making. Thanks to data expressed in concepts, we mask unwanted $k$-th features across all training samples as follows: $\psi_i^{k}=\psi_{i}\odot (e - e_k)$, where $e \in \mathbb{R}^{d_\sfPsi}$ denotes all-one-vectors, and $\odot$ represents the Hadamard product. We train the classifier $P$ with $\psi_{train}^k=\{E(x_i)\odot (e-e_k) \ | \ x_i \in \mathcal{D}_{train} \}$. As a result, Img2Tabs exclude the specific unwanted concepts from decision-making.

\section{Evaluation and Discussion}
\quad In this section, we provide three evaluations as follows: 1) we demonstrate concept-based explanation by Img2Tab, and 2) we measure the top-1 accuracy of Img2Tab to compare with CNN and deep feature learning baselines. 3) We show how Img2Tab debugging affects important concepts and top-1 accuracy. Img2Tab employs the XGBoost classifier for all evaluations. The datasets used in the subsequent experiments are the CelebA (face recognition) and MNIST (digit recognition). Note that GAN inversion networks of the face domain are pre-trained by the FFHQ dataset for generalized concept learning. Refer to the Supplementary for more implementation details.

\subsection{Img2Tab Explanations}
\quad \textbf{Visualizing important concepts} We begin by visualizing important concepts extracted by Img2Tabs emphasizing that Img2Tab automatically discovers them without any manual selections using $W_k$ metrics. As exhibited in Figure 6, important concepts automatically discovered by the Img2Tab are diverse, rich, well-disentangled, and human-understandable in various samples. By common sense, we observe that they are highly relevant to the class and human-understandable concepts.
\begin{figure}[!t]
  \centering
  \setlength{\belowcaptionskip}{-10pt}
  \includegraphics[width=0.95\linewidth]{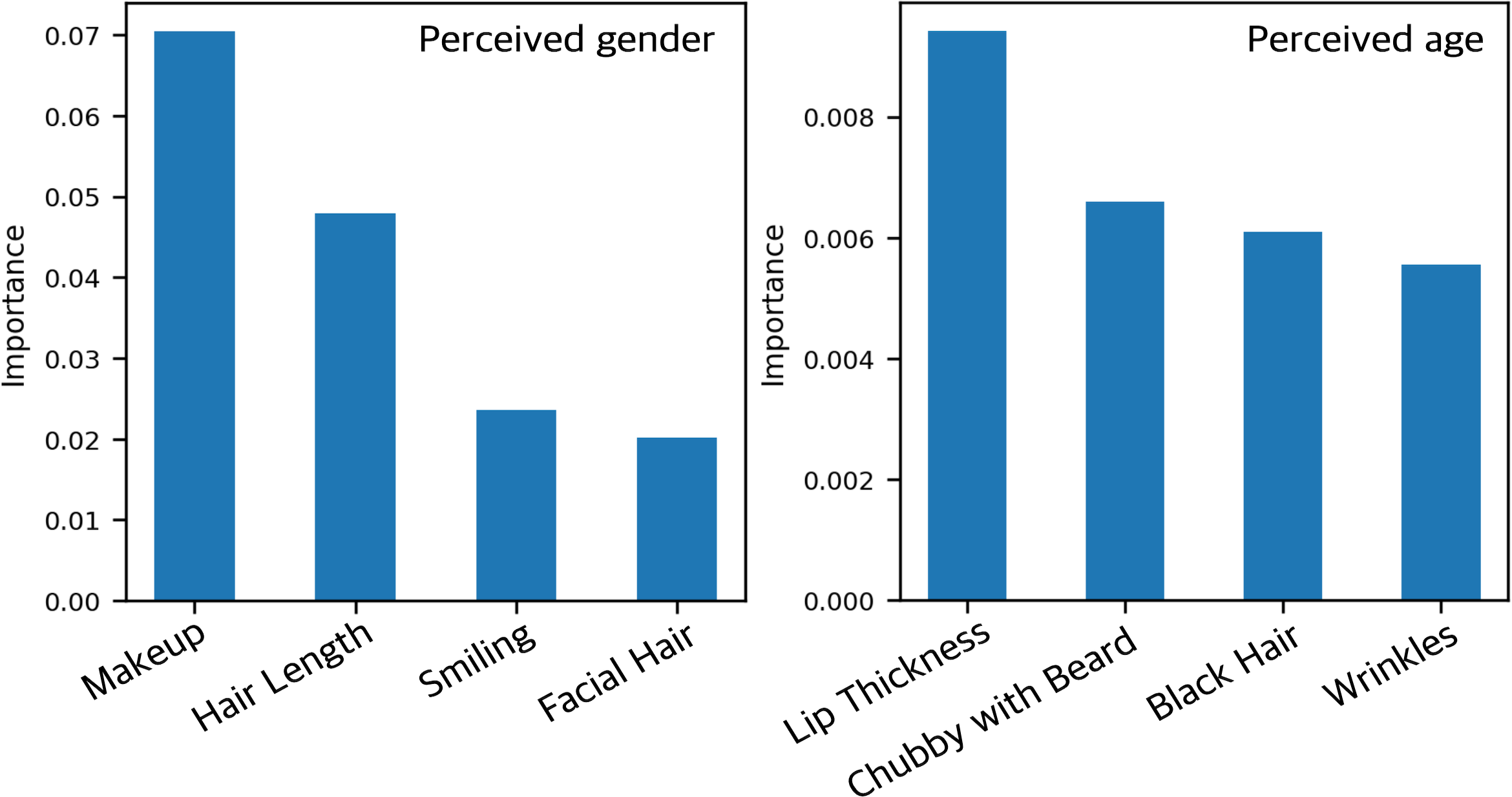}
  \caption{Importance plots for the top 4 important concepts discovered by Img2Tab for perceived gender classification and age classification.}
   \label{fig:onecol}
\end{figure}

\quad\textbf{Identifying concept importance}
Figure 7 displays the importance of each important concept. The Img2Tab framework explains its inference in that ``Makeup," ``Hair Length," ``Smiling," and ``Facial Hair" are of ranked importance (descending order) for the perceived gender classification. We take exception with ``Smiling" as a concept, further discussed in Section 5.3. For perceived age classification, the Img2Tab decision-making primarily leverages ``Lip Thickness," ``Chubby with Beard," ``Black hair," and ``Wrinkles." The feature importance explanation epitomizes the most intuitive method, allowing users to identify the most influential concepts in the prediction at a glance.

\begin{table}[b!]
    \begin{tabular}{lcccc}
        \toprule
        Classifiers & Gender & Age & Eyeglasses & Bangs \\
        \midrule
        ResNet-50       & 97.04 & 87.27 & 99.01 & 95.39 \\
        Inception-v4    & 97.24 & 87.05 & 99.00 & 95.17 \\
        MobileNet-v3    & 97.10 & 87.25 & 98.97 & 95.33 \\
        DenseNet-121    & 97.53 & 87.49 & 99.06 & 95.24 \\        
        EfficientNet-b4 & 98.10 & 88.01 & 99.43 & 96.02 \\
        \bottomrule
        Img2Tab(Ours)   & 97.85 & 87.44 & 99.23 & 94.88 \\
        \hline
    \end{tabular}
    \caption{Top-1 accuracy[\%] comparison for CelebA attributes}
    \label{tab:booktabs}
\end{table}

\subsection{Performance Evaluations}
\quad In this section, we compare Img2Tabs to CNN baselines on top-1 accuracy, showing demonstrating that Img2Tabs achieve comparable performance to black-box baselines.

\textbf{Img2Tab vs CNN} Table 1 reports the top-1 accuracy of Img2Tab and CNN baselines for CelebA attribute classifications. Every CNN baseline is initialized with ``ImageNet" pre-training. Our Img2Tab outperforms ResNet-50 \cite{he2016deep}, Inception-v4 \cite{szegedy2017inception}, and MobileNet-v3 \cite{howard2019searching} with 0.85\% in maximum except `Bangs' classification. DenseNet-121 achieves top-1 accuracy nearly on par with our Img2Tab. Meanwhile, EfficientNet-b4 \cite{tan2019efficientnet} outperforms our Img2Tab slightly with 0.6\% on average. This suggests that the discriminative capability of Img2Tab is comparable to CNN baselines. Thus, Img2Tab frameworks possess tremendous potential to compete with other CNN-based models while retaining explainability.

\textbf{Img2Tab vs Deep feature learning} As Img2Tab uses GAN inversion for feature(concept) learning, we compare deep feature learning methods to Img2Tab in the MNIST digit recognition task. Note that the baseline results in Table 2 adhere to the GH-Feat paper \cite{xu2021generative} for this evaluation. Our Img2Tab accuracy surpasses AEs \cite{hinton2006reducing}, BiGAN \cite{donahue2016adversarial} and ALAE \cite{pidhorskyi2020adversarial} by 0.81\%, 0.87\%, 1.1\%, and 0.63\% margin, respectively. Slightly lower accuracy compared to the GH-FEAT appears at the cost of explainability with Img2Tab, as the decision-making of GH-Feat \cite{xu2021generative} is not explainable.

\begin{table}[t!]
    \centering
    \begin{tabular}{lc}
        \toprule
         Methods            & Digit recognition \\
        \midrule
        AE(\mymath{l_1})    & 97.43 \\
        AE(\mymath{l_2})    & 97.37 \\
        BiGAN               & 97.14 \\
        ALAE                & 97.61 \\
        GH-Feat             & 99.06 \\ 
        \bottomrule
        Img2Tab(Ours)       & 98.24 \\ \hline
    \end{tabular}
    \caption{Top-1 accuracy[\%] comparison for MNIST digit recognition}
    \label{tab:booktabs}
\end{table}

\begin{figure}[H]
  \centering
  \includegraphics[width=1.0\linewidth]{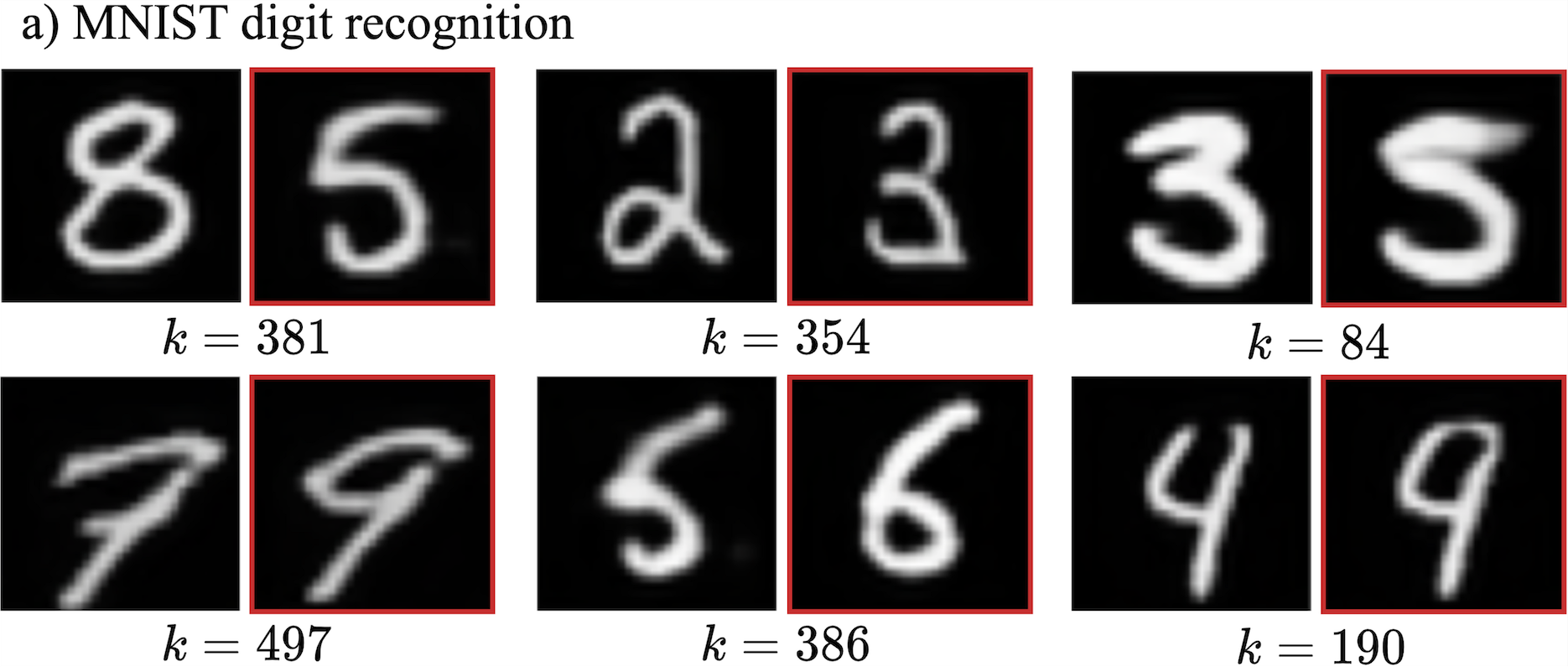}
  \caption{Important concepts automatically discovered by Img2Tab for MNIST digit recognition. For each image pair, the left captures the source image, and the red-outlined represents $G(x_{i,\pm k})$ of the corresponding image.}
   \label{fig:onecol}
\end{figure}

\subsection{Debugging Img2Tab Classifier}
\quad In previous evaluations, Img2Tab erroneously includes concepts of ``Smiling" and ``Makeup" for classifying gender. By common sense, ``Smiling" does not serve as a reasonable concept when classifying gender. But Img2Tab considers the concept important because females in CelebA samples smile more often than males, which invites bias. Also ``Makeup" is not an inborn characteristic of females leading to unfair decisions. We debug the Img2Tab classifier by masking ``Makeup" and ``Smiling" at the concept level(i.e., not by masking pixels in an image) to ensure unbiased and fair decision-making. Figure 9 depicts the importance of perceived gender classification before and after debugging. Accordingly, the importance of ``Makeup" and ``Smiling" becomes zero. The debugged model does not include any information about ``Makeup" and ``Smiling" concepts, which yields fair and unbiased decision-making.

To check whether masking concepts reduces the accuracy, we debug Img2Tab with each top-4 important concept masked: $\psi_i^{4319}$(``Makeup"), $\psi_i^{503}$(``Hair Length"), $\psi_i^{4440}$(``Smiling"), and $\psi_i^{4614}$(``Facial Hair''). The accuracy results are 97.90\%, 97.84\%, 97.83\%, and 97.95\%; accuracy barely changes compared to the baseline(97.80\%). Img2Tabs effectively discover alternative class-relevant concepts, despite one concept being masked, which captures the flexibility of Img2Tab decision-making.

\begin{figure}[t!]
  \centering
  \includegraphics[width=1.0\linewidth]{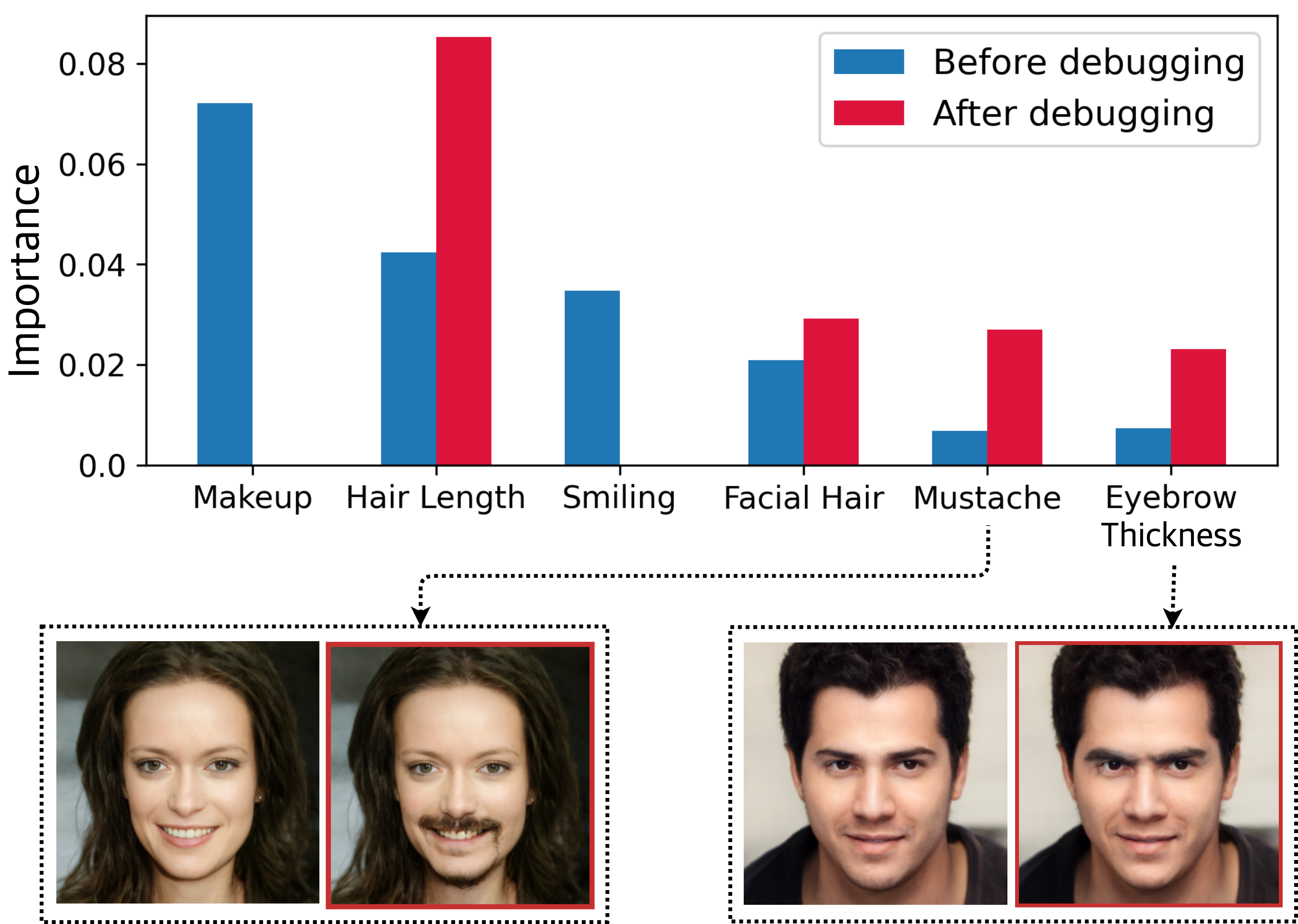}
  \caption{A plot comparing importance changes after masking unwanted concepts for unbiased and fair decision-making. We mask `Makeup' and `Smiling' concepts.}
   \label{fig:onecol}
\end{figure}
\section{Conclusion}
\quad We proposed a novel method using the Wasserstein-1 metric to measure class-relevancy and interpretability simultaneously from StyleGAN features. By this method, we evaluate that training XGBoost within StyleGAN features enables Img2Tabs to produce explainable concept-based image classifications. Img2Tab predicts image labels and explains its prediction through importance scores and visualization at the concept level. In addition, users can easily debug the Img2Tab classifier to ensure unbiased and fair decision-making. Img2Tabs gains several advantages over existing concept-based explanations whilst not encountering significant accuracy reduction compared to the deep net baselines. 

This paper bridges the gap between generative models and tabular classifiers to harness the explainability of the latter for image classifications. Since GAN inversion networks and tabular classifiers are not trained jointly, there is room to enhance the classification performance and quality of the feature space, with the advent of more sophisticated techniques from each research domain.

\bibliographystyle{named}
\bibliography{reference}

\end{document}